\documentclass[10pt,twocolumn,letterpaper]{article}
\usepackage{cvpr}
\usepackage{times}
\usepackage{epsfig}
\usepackage{graphicx}
\usepackage{amsmath}
\usepackage{amssymb}
\usepackage{mathtools}
\usepackage{subfigure}
\usepackage{color}
\usepackage{soul}
\usepackage[linesnumbered,vlined,ruled]{algorithm2e}
\usepackage{algorithmic}
\usepackage{enumitem}
\usepackage{url}
\usepackage{bm}
\usepackage[T1]{fontenc} 
\usepackage{xspace}
\usepackage{caption}
\usepackage{import}
\usepackage{gensymb}
\usepackage{epstopdf}

\newcommand{\qed}{\nobreak \ifvmode \relax \else
      \ifdim\lastskip<1.5em \hskip-\lastskip
      \hskip1.5em plus0em minus0.5em \fi \nobreak
      \vrule height0.75em width0.5em depth0.25em\fi}

\newcommand{\bx}{\mathbf{x}}
\newcommand{\cX}{\mathcal{X}}
\newcommand{\cI}{\mathcal{I}}

\newcommand{\cD}{\mathcal{D}}

\newcommand{\cF}{\mathcal{F}}
\newcommand{\cP}{\mathcal{P}}
\newcommand{\cL}{\mathcal{L}}
\newcommand{\cH}{\mathcal{H}}

\newcommand{\bbf}{\mathbf{f}}

\newcommand{\bX}{\mathbf{X}}

\newcommand{\Mod}[1]{\ \mathrm{mod}\ #1}

\usepackage{color}
\usepackage{xcolor}
 \SetKwRepeat{Do}{do}{while}

\usepackage[pagebackref=true,breaklinks=true,letterpaper=true,colorlinks=true,bookmarks=false]{hyperref}
\definecolor{orange}{rgb}{0,0.9,0.2}
\hypersetup{
linkcolor=red,
anchorcolor=black,
citecolor=orange,
filecolor=cyan,
menucolor=red,
runcolor=cyan,
urlcolor=magenta,
}

\graphicspath{{images/}}
\usepackage{caption}
\cvprfinalcopy 

\def\cvprPaperID{453} 

\begin{document}

\title{SPP-Net: Deep Absolute Pose Regression with Synthetic Views}
\author{Pulak Purkait, Cheng Zhao, Christopher Zach\\
Toshiba Research Europe, 
Cambridge, UK\\
{\tt\small pulak.isi@gmail.com}
}

\maketitle

\begin{abstract}
  Image based localization is one of the important problems in computer vision
  due to its wide applicability in robotics, augmented reality, and autonomous
  systems. There is a rich set of methods described in the literature how to
  geometrically register a 2D image w.r.t.\ a 3D model. Recently, methods
  based on deep (and convolutional) feedforward networks (CNNs) became popular
  for pose regression. However, these CNN-based methods are still less
  accurate than geometry based methods despite being fast and memory
  efficient. In this work we design a deep neural network architecture based
  on sparse feature descriptors to estimate the absolute pose of an image. Our
  choice of using sparse feature descriptors has two major advantages: first,
  our network is significantly smaller than the CNNs proposed in the
  literature for this task---thereby making our approach more efficient and
  scalable. Second---and more importantly---, usage of sparse features allows to augment the training
  data with synthetic viewpoints, which leads to substantial improvements in
  the generalization performance to unseen poses. Thus, our proposed method
  aims to combine the best of the two worlds---feature-based localization and
  CNN-based pose regression--to achieve state-of-the-art performance in the
  absolute pose estimation. A detailed analysis of the proposed architecture
  and a rigorous evaluation on the existing datasets are provided to support
  our method.

\end{abstract}

\section{Introduction}

Image localization is the task of accurately estimating the location and
orientation of an image with respect to a global map and has been studied
extensively in robotics and computer vision. In this work we consider
the more specific setting of estimating the perspective pose of an image with
respect to a given 3D model, in particular 3D point clouds. Traditionally,
this problem has been tackled either by direct 2D-3D matching
(e.g.~\cite{li2010location,sattler2017efficient}) or by inserting an image
retrieval stage to narrow down the search space
(e.g.~\cite{zhang2006image,schindler2007cityscale,irschara2009structure}). PoseNet~\cite{kendall2015posenet}
and related
approaches~\cite{weyand2016planet,kendall2017geometric,walch2017image}
demonstrated that deep learning methods---which have shown excellent
performance in numerous classification and regression problems---are also able
to estimate camera poses directly from input images.

\begin{figure}
\includegraphics[width=0.49\textwidth]{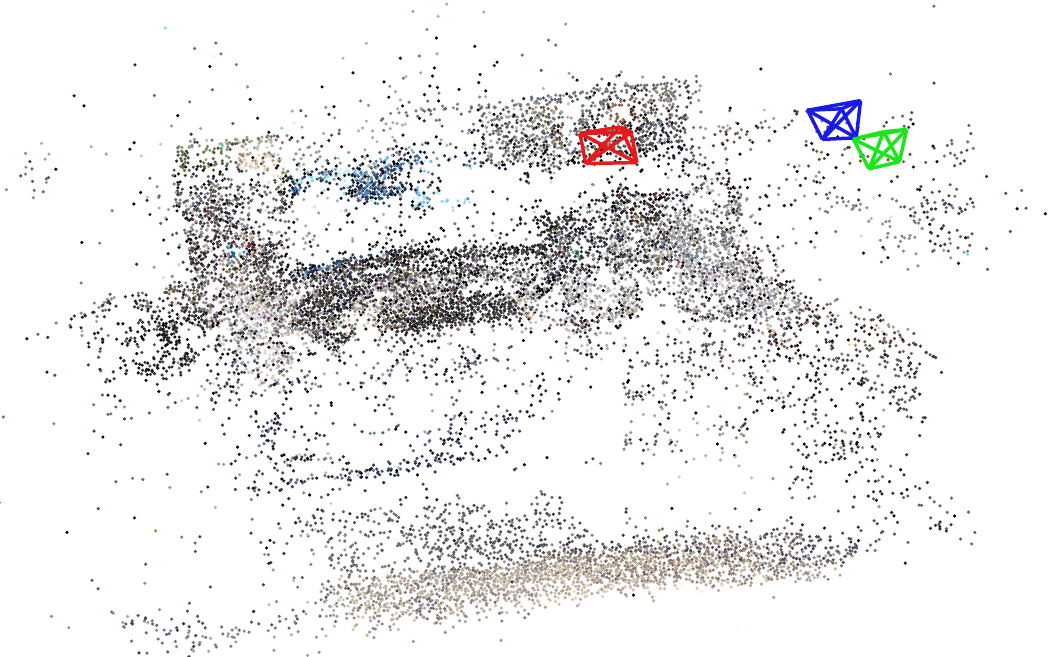}
\caption{An example of 6DOF pose estimation results on heads sequence of the Seven Scenes dataset \cite{glocker2013real} where PoseNet~\cite{kendall2015posenet} fails  to predict accurate pose (marked by red, positional error $=0.31$m and angular error $=27.4\degree$) whereas the proposed SPP-Net predicts a pose (marked by blue, positional error $=0.06$m and angular error $=2.18\degree$) close to the ground truth (marked by green). }
\label{fig:token}
\end{figure}

Despite the good performance of PoseNet and subsequent architectures for pose
regression from
images~\cite{kendall2015posenet,kendall2017geometric,walch2017image}, we
believe that PoseNet-like methods are fundamentally limited in the following
ways:
\begin{enumerate}[leftmargin=1em,itemsep=1pt,parsep=1pt,topsep=1pt]
\item Forward regression architectures such as deep neural networks have no
  built-in reasoning about geometry and most likely do not extract an
  ``understanding'' of the underlying geometric concepts during the training
  phase. Consequently, we postulate (and empirically validate) that
  PoseNet-like approaches suffer from poor extrapolation ability to unseen
  poses that are significantly different from the ones in the training set. In
  many application settings the distributions of training poses and test poses
  can differ substantially: training images (and poses) might be chosen such
  that structure-from-motion computation to obtain a 3D model is made easier,
  whereas test poses may be arbitrarily distributed within the maneuverable
  space. Hence, pose regression typically faces a severe domain adaptation
  problem in general.
\item A lot of computation (and trainable parameters) in PoseNet-like
  architectures goes into the feature extraction stage, which uses rather
  heavy-weight CNNs such as VGG-Net~\cite{simonyan2015verydeep} or
  GoogleNet~\cite{szegedy2015going}. In light of empirical evidence that
  gold-standard feature descriptors (such as SIFT~\cite{lowe2004distinctive}) in
  general have better accuracy in pose estimation than CNN-based approaches,
  we conjecture that heavy-weight dense feature extraction via CNNs is not
  necessary for this task. Hence, the networks used in our approach are
  significantly smaller and faster to train than existing CNN-based solutions
  for pose regression. Our choice to rely only on sparse features will be also
  very beneficial to address the above-mentioned domain adaptation problem.
\end{enumerate}
The goal in this work is to close the gap between explicit
correspondence-based methods and deep learning methods for pose
regression. The general advantages of using deep learning for pose regression
are the benefits of end-to-end training, the reduced memory requirements
($\approx\!100\,$MB for the network instead of several GB for a typical 3D point
cloud database), and real-time performance (less than $50\,$ms for pose
estimation per image).

\noindent {\bf  Our contributions can be summarized as follows: }
\begin{itemize}[leftmargin=1em,itemsep=1pt,parsep=1pt,topsep=1pt]
\item We address the domain adaptation problem in pose estimation by
  augmenting the training set with synthetically generated training images and
  poses. In the spirit of~\cite{irschara2009structure} these synthetic poses
  may cover regions in pose space not available in the training data.
\item Our choice of sparse features as input to the DNN implies, that we do
  not have to generate realistic RGB images for synthetic poses, but only have
  to predict realistic sets of features and associated descriptors.
\item We refine the method in~\cite{irschara2009structure} to generate synthetic images leveraging the 3D map and feature correspondences.
\item We propose a DNN architecture based on an ensemble of spatial pyramid
  max-pooling units~\cite{he2014spatial} for pose regression. This network can
  be trained (from scratch) on those real+synthetic datasets without
  pretraining and is significantly smaller than PoseNet-like networks reported
  in the literature.
\item We evaluate our proposed DNN on standard datasets for pose estimation,
  and we put a particular focus on the generalization performance to unseen
  test poses. We demonstrate that our method reduces the performance gap
  between the training based and direct feature based methods and produces
  state-of-the-art results among training based methods on benchmark datasets.
\end{itemize}
Thus, in this work we show that relatively light-weight pose regression
network can achieve state-of-the-art results in CNN-based pose regression, and
we demonstrate that adding synthesized data to the training set substantially
improves its generalization ability to novel poses.

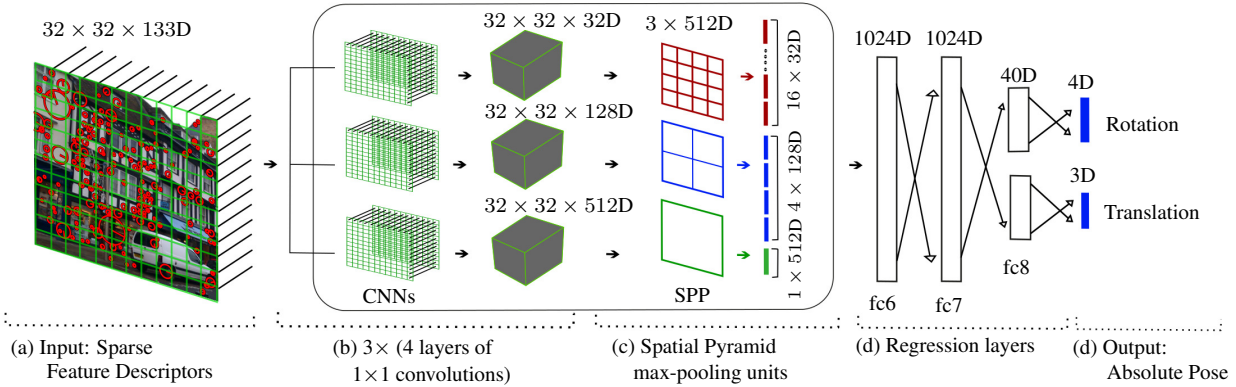
\begin{figure*}
\centering \footnotesize \hspace{-1cm}
\def\svgwidth{0.98\textwidth}
\begingroup%
  \makeatletter%
  \providecommand\color[2][]{%
    \errmessage{(Inkscape) Color is used for the text in Inkscape, but the package 'color.sty' is not loaded}%
    \renewcommand\color[2][]{}%
  }%
  \providecommand\transparent[1]{%
    \errmessage{(Inkscape) Transparency is used (non-zero) for the text in Inkscape, but the package 'transparent.sty' is not loaded}%
    \renewcommand\transparent[1]{}%
  }%
  \providecommand\rotatebox[2]{#2}%
  \ifx\svgwidth\undefined%
    \setlength{\unitlength}{550.78933324bp}%
    \ifx\svgscale\undefined%
      \relax%
    \else%
      \setlength{\unitlength}{\unitlength * \real{\svgscale}}%
    \fi%
  \else%
    \setlength{\unitlength}{\svgwidth}%
  \fi%
  \global\let\svgwidth\undefined%
  \global\let\svgscale\undefined%
  \makeatother%
  \begin{picture}(1,0.38317731)%
    \put(0,0){\includegraphics[width=\unitlength]{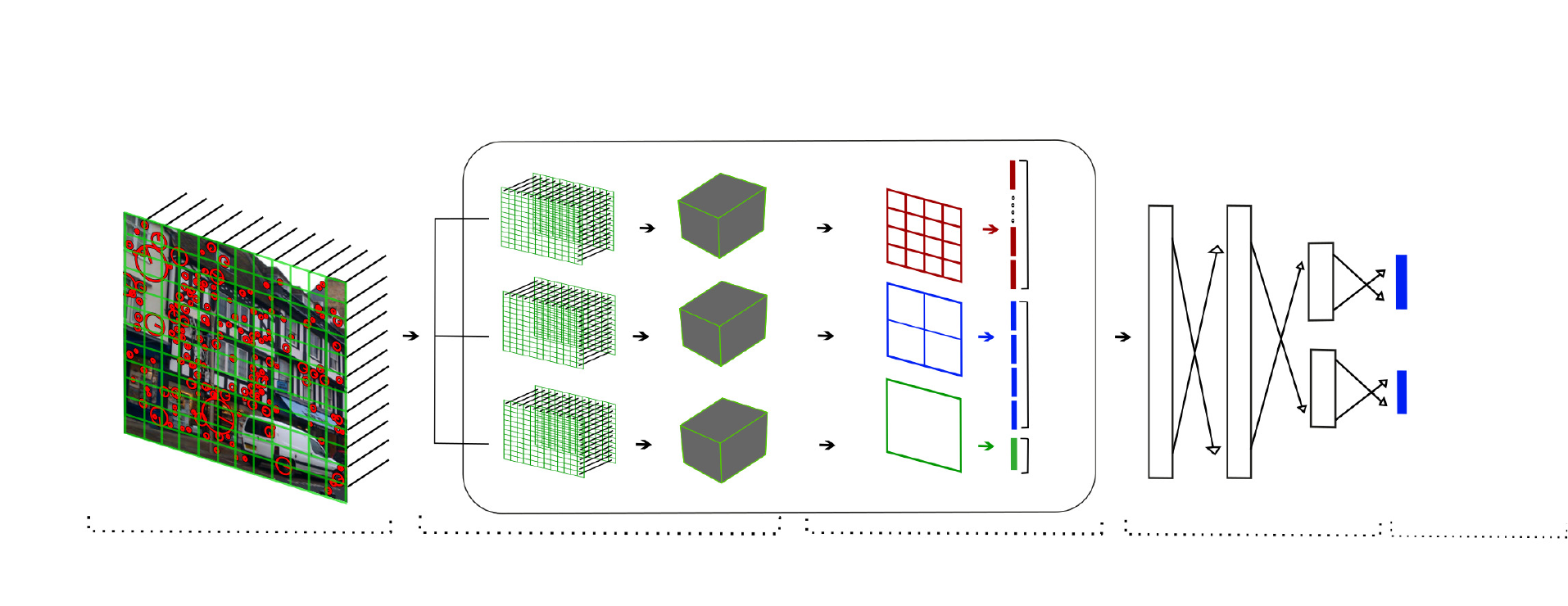}}%
    \put(0.27282328,-0.61146531){\color[rgb]{0,0,0}\makebox(0,0)[lt]{\begin{minipage}{0.01452461\unitlength}\raggedright \end{minipage}}}%
    \put(0.49579829,0.4160763){\color[rgb]{0,0,0}\makebox(0,0)[lb]{\smash{}}}%
    \put(0.05601658,-0.14885032){\color[rgb]{0,0,0}\makebox(0,0)[lt]{\begin{minipage}{1.11694247\unitlength}\raggedright \end{minipage}}}%
    \put(0.08070841,-0.11399126){\color[rgb]{0,0,0}\makebox(0,0)[lt]{\begin{minipage}{0.87002411\unitlength}\raggedright \end{minipage}}}%
    \put(0.84343055,0.05994974){\color[rgb]{0,0,0}\makebox(0,0)[lb]{\smash{}}}%
    \put(0.26580439,0.0874219){\color[rgb]{0,0,0}\makebox(0,0)[lb]{\smash{}}}%
    \put(0.70744531,0.19667652){\color[rgb]{0,0,0}\makebox(0,0)[lt]{\begin{minipage}{0.03177258\unitlength}\raggedright \end{minipage}}}%
    \put(0.33334085,0.06396716){\color[rgb]{0,0,0}\makebox(0,0)[lb]{\smash{CNNs}}}%
    \put(0.57546964,0.06519186){\color[rgb]{0,0,0}\makebox(0,0)[lb]{\smash{SPP}}}%
    \put(0.72651946,0.05914434){\color[rgb]{0,0,0}\makebox(0,0)[lb]{\smash{fc6}}}%
    \put(0.77824698,0.05711776){\color[rgb]{0,0,0}\makebox(0,0)[lb]{\smash{fc7}}}%
    \put(0.83074466,0.08702288){\color[rgb]{0,0,0}\makebox(0,0)[lb]{\smash{fc8}}}%
    \put(0.91073621,0.19793247){\color[rgb]{0,0,0}\makebox(0,0)[lb]{\smash{Rotation }}}%
    \put(0.90921079,0.12974787){\color[rgb]{0,0,0}\makebox(0,0)[lb]{\smash{Translation}}}%
    \put(0.88129897,0.23061725){\color[rgb]{0,0,0}\makebox(0,0)[lb]{\smash{4D}}}%
    \put(0.88264292,0.15803879){\color[rgb]{0,0,0}\makebox(0,0)[lb]{\smash{3D}}}%
    \put(0.55277959,0.27681511){\color[rgb]{0,0,0}\makebox(0,0)[lb]{\smash{$3\times512$D}}}%
    \put(0.71481081,0.26510105){\color[rgb]{0,0,0}\makebox(0,0)[lb]{\smash{1024D}}}%
    \put(0.77021234,0.26506298){\color[rgb]{0,0,0}\makebox(0,0)[lb]{\smash{1024D}}}%
    \put(0.82892323,0.23634343){\color[rgb]{0,0,0}\makebox(0,0)[lb]{\smash{40D}}}%
    \put(0.08439085,0.27161195){\color[rgb]{0,0,0}\makebox(0,0)[lb]{\smash{$32\times32\times133$D}}}%
    \put(0.05976863,0.02508935){\color[rgb]{0,0,0}\makebox(0,0)[lb]{\smash{(a) Input: Sparse}}}%
    \put(0.42800979,0.20817956){\color[rgb]{0,0,0}\makebox(0,0)[lb]{\smash{$32\times32\times128$D}}}%
    \put(0.67243787,0.21335639){\color[rgb]{0,0,0}\rotatebox{89.21801152}{\makebox(0,0)[lb]{\smash{\scriptsize  $16\times32$D}}}}%
    \put(0.673039,0.13503039){\color[rgb]{0,0,0}\rotatebox{89.20402481}{\makebox(0,0)[lb]{\smash{\scriptsize $4\times128$D}}}}%
    \put(0.67324482,0.06638516){\color[rgb]{0,0,0}\rotatebox{89.20402481}{\makebox(0,0)[lb]{\smash{\scriptsize $1\times 512$D}}}}%
    \put(0.31005223,0.02434359){\color[rgb]{0,0,0}\makebox(0,0)[lb]{\smash{(b) 3$\times$ (4 layers of}}}%
    \put(0.52672549,0.02555701){\color[rgb]{0,0,0}\makebox(0,0)[lb]{\smash{(c) Spatial Pyramid}}}%
    \put(0.71625033,0.02587642){\color[rgb]{0,0,0}\makebox(0,0)[lb]{\smash{(d) Regression layers }}}%
    \put(0.88484564,0.02473397){\color[rgb]{0,0,0}\makebox(0,0)[lb]{\smash{(d) Output:}}}%
    \put(0.427004,0.27893066){\color[rgb]{0,0,0}\makebox(0,0)[lb]{\smash{$32\times32\times32$D}}}%
    \put(0.4278579,0.13509535){\color[rgb]{0,0,0}\makebox(0,0)[lb]{\smash{$32\times32\times512$D}}}%
    \put(0.08785189,0.00583227){\color[rgb]{0,0,0}\makebox(0,0)[lb]{\smash{Feature Descriptors}}}%
    \put(0.32489623,0.00388292){\color[rgb]{0,0,0}\makebox(0,0)[lb]{\smash{1$\times$1 convolutions)}}}%
    \put(0.54357547,0.00549753){\color[rgb]{0,0,0}\makebox(0,0)[lb]{\smash{max-pooling units  }}}%
    \put(0.91132415,0.00547689){\color[rgb]{0,0,0}\makebox(0,0)[lb]{\smash{Absolute Pose}}}%
    \put(0.60531915,-0.12842509){\color[rgb]{0,0,0}\makebox(0,0)[lb]{\smash{}}}%
  \end{picture}%
\endgroup%
\caption{Proposed SPP-Net for absolute pose regression takes sparse feature points as input and predicts the absolute pose. }
\label{fig:proposedModel}
\end{figure*}

\section{Related Work} 

\noindent {\bf Localization from 3D structure}
Localization and estimation of the camera pose with respect to a 3D
environment, that is known in advance, is a well-studied problem in the
field. A number of existing works
(e.g.~\cite{li2010location,sattler2011fast,li2016worldwide,sattler2017efficient})
draws its inspiration from the related problem of detecting loop closures in
visual SLAM~\cite{newman2005slam,cummins2008fabmap,angeli2008fast} by
explicitly searching for 2D-3D correspondences (and subsequently determining the pose by a
robust perspective $N$-point algorithm). If the underlying 3D point cloud is
very large, then adding an intermediate, bag-of-features based step to quickly
identify relevant subsets of the point cloud can be highly beneficial to reduce the
computational
costs~\cite{zhang2006image,schindler2007cityscale,irschara2009structure}. One
problem with methods using a 3D point cloud is a somewhat limited scalability:
works such as~\cite{li2016worldwide} use relatively large point clouds
containing many millions of 3D points, but ultimately cover only a tiny
fraction of the world.


\noindent {\bf Localization via image retrieval}
One way to make location recognition more scalable is by casting it as an
image retrieval task. In image retrieval it is relatively well understood how
to index large image collections for efficient search, which can be based on
bag-of-features
(e.g.~\cite{nister2006scalable,philbin2007object,jegou2010improving}) or more
general global image descriptors (such
as~\cite{torralba2008small,weiss2009spectral,jegou2008hamming,perronnin2010large}). Without
an underlying 3D model, location (or place / landmark) recognition via image
retrieval can only provide an approximate estimate for the pose and will not
determine all necessary 6 d.o.f.\ accurately.


\noindent {\bf Learning based algorithms}
To our knowledge the first attempt to use machine learning techniques
specifically for pose estimation is~\cite{shotton2013scene}, where a random
forest regressor is used to (densely) predict initial 2D-3D correspondences
from image appearance. The resulting pose is obtained by subsequent robust
estimation. PoseNet~\cite{kendall2015posenet} is probably the first approach
aiming for end-to-end, CNN-based pose regression from an input image. The
network (based on a pre-trained GoogleNet) has to extract higher-level and
invariant properties of an image in order to accurately predict the pose
parameters. \cite{walch2017image} enhance in the regression layers by
utilizing LSTMs, and~\cite{kendall2017geometric} improve upon the original
PoseNet by leveraging different loss functions,
and~\cite{clement2017train,laskar2017camera,melekhov2017image} are recent
developments in direct CNN-based pose estimation.

Since none of the above deep learning architecture for pose regression has an
intrinsic ``understanding'' of the geometric concepts behind pose estimation,
the generalization performance to new poses that are significantly different
to the training poses is expected to be poor. Note that ``geometric'' methods
for pose estimation will suffer to some extent from the same problem (e.g.\
due to limited view-point invariance of feature descriptors), but will in
general much better extrapolate to novel poses. This work aims improve the
generalization ability of CNN-based pose regression by using an architecture
that is capable to learn from synthesized training examples.



\section{Spatial pyramid pose net}\label{sec:overview}
In the following we describe our proposed DNN architecture, which we term ``spatial pyramid pose net'' or SPP-Net.
SPP-Net takes a set of sparse feature descriptors as input and estimates 6 d.o.f.\ camera pose. The input descriptors undergo a number of $1\times 1$ convolutions/ReLU layers, followed by an ensemble of multiple parallel max-pooling layers, which are succeeded by three fully connected pose regression layers. The proposed SPP-Net is a light-weight, fast, and it performs analogously with the original PoseNet~\cite{kendall2015posenet}. Moreover, the proposed architecture has an additional advantage that it can be further trained on augmented images generated from the reconstructed 3D map. Being trained on such synthetic poses, it produces state-of-the-art results on benchmark datasets. In the following subsections, we describe the proposed network in detail and in section~\ref{sec:mining}, we present our approach for simulating augmented views.

\paragraph{Feature processing}
Given a set of input keypoints $\cX=\{\bx_1, \bx_2,\ldots \bx_N\}$, the task is to estimate the pose of the camera $(R, T)$ with respect to the global coordinate frame. The $i$-th keypoint $\bx_i$ is described by its pixel coordinates $(p_i, q_i)$, scale $s_i$, orientation $\theta_i$, and a feature descriptor $\bbf_i$ of dimensionality $D$.
The number of of sparse features extracted from input images varies, and there
are two complementary approaches to facilitate the use of a CNN on sparse
features: the method used in~\cite{dosovitskiy2016inverting} embeds the sparse
features into a dense grid (and has to handle feature collisions and empty
locations), and PointNet~\cite{qi2016pointnet} (and
similarly~\cite{tolias2015particular}) processes each input feature
independently and uses max-pooling to symmetrize the network output. Our
approach combined elements of both by using spatial binning, indepdent feature
processing and global max-pooling:
we arrange the set of keypoints on a 2D regular grid based on the pixel locations $(p_i , q_i)$ and split the input image of size $W\times H$ into
$d_1\times d_2$ cells (where each cell is of size $W/d_1 \times H/d_2$ pixels). 
If a cell occupies multiple features, we select a feature randomly and obtain a $(D+5)$-dimensional vector $\big(\bbf_i, p_i\Mod{d_1}, q_i\Mod{d_2}, \sin \theta_i, \cos \theta_i, \log (1 + s_i)\big)$ corresponding to a keypoint $\bx_i$.
This results a spatially organized array of at most $d_1 \times d_2$ feature descriptors (with $D+5$ dimensions), which is the input to the network. Empty cells are represented by a zero feature vector.
In all of our experiments, we used $d_1 = d_2 = 32$ for the (outdoor) datasets of large images and $d_1 = d_2 = 16$ for the (indoor) datasets of small images.

The rationale behind our spatial binning approach is to reduce the amount of processing and to balance the spatial distribution of features across the image.

\paragraph{Network Architecture}
As shown in Figure~\ref{fig:proposedModel}, the proposed network consists of an array of CNN subnets, an ensemble layer of max-pooling units at different scales and two fully connected layers followed by the output pose regression layer. At each scale, a CNN feature descriptors is fed to the ensemble layer of multiple maxpooling units [Fig.~\ref{fig:proposedModel}(b)]. A CNN consists of $4$ convolution layers of size $1\times 1$ of dimensionally $D^\prime_s$ which are followed by relu activation and batch normalization. Thus, the set of $d_1 \times d_2$, $(D+5)$-dimensional input descriptors is fed into the CNNs at multiple scales, each of which produces feature map of size $ d_1 \times d_2 \times D^\prime_s$. Note that the number of feature descriptors is unaltered during the convolution layers. Experimentally we have found that the chosen $1\times 1$ convolutions with stride $1\times 1$ performs better than larger convolutions. In all of our experiments, we utilize SIFT descriptors of size $D=128$ and the dimension of the CNN feature map $D^\prime_s$ at level $s$ is chosen to be $D^\prime_s = 512/2^{2s}$. 

Inspired by spatial pyramid pooling~\cite{he2014spatial}, in SPP-Net we concatenate the outputs of the individual max-pooling layers before reaching the final fully connected regression layers.
We use parallel max-pooling layers at several resolutions:
at the lowest level of the ensemble layer has $D_0'$ global max-pooling units (each taking $d_1 \times d_2$ inputs), and at the $s$th level it has $2^{2s} \times D_s'$ max-pooling units (with a receptive field of size $d_1/(2^s) \times d_2/(2^s)$).
The response of all the max-pooling units are then concatenated to get a fixed length feature vector of size $\sum_s2^{2s}\times 512/2^{2s} = 512\times(s+1)$. In all of our experiments, we have chosen a fixed level $s=2$ of max-pooling unites. Thus, the number of output feature channel of the ensemble layer is $D^\prime = 1536$. The feature channels are then fed into two subsequent fully connected layers (fc6 and fc7 of Fig.~\ref{fig:proposedModel}) of size $1024$. We also incorporate dropout strategy for the fully connected layers with probability $0.5$. The fully connected layers are then split into two separate parts, each of dimension $40$ to estimate $3$-dimensional translation and $4$-dimensional quaternion separately. 

The number of parameters and the operations used in different layers are demonstrated in Table~\ref{tab:paramnum}. A comparison among different architectures can also be found in Table~\ref{tab:paramcomp}. 
\begin{table}[!bht]\setlength{\tabcolsep}{4pt}
\scriptsize
\begin{center}
  \begin{tabular}{ccccc} \hline 
  	type~/~depth	& 	patch~size~/~stride & output	&	$\#$params	& $\#$ FLOPs \\ \hline \hline   
  	conv0/1 & $1\times 1/1$ & $32\times32\times128$ & $17$K & $17$M \\ 
  conv0/2 & $1\times 1/1$ & $32\times32\times256$ & $32.7$K & $32.7$M \\ 
  conv0/3 & $1\times 1/1$ & $32\times32\times256$ & $65.5$K & $65.5$M \\ 
  conv0/4 & $1\times 1/1$ & $32\times32\times512$ & $131$K & $131$M \\ \hline 
 
  conv1/1 & $1\times 1/1$ & $32\times32\times128$ & $17$K & $17$M \\ 
  conv1/2 & $1\times 1/1$ & $32\times32\times128$ & $16.4$K  & $16.4$M \\ 
  conv1/3 & $1\times 1/1$ & $32\times32\times128$ & $16.4$K & $16.4$M \\ 
  conv1/4 & $1\times 1/1$ & $32\times32\times128$ & $16.4$K & $16.4$M \\ \hline 
  
  conv2/1 & $1\times 1/1$ & $32\times32\times128$ & $17$K & $17$M \\ 
  conv2/2 & $1\times 1/1$ & $32\times32\times64$ & $8.3$K & $8.3$M \\ 
  conv2/3 & $1\times 1/1$ & $32\times32\times64$ & $4.1$K  & $4.1$M \\ 
  conv2/4 & $1\times 1/1$ & $32\times32\times32$ & $2$K & $2$M \\ \hline 
  
  max-pool0/5 & $32\times32/32$ & $1\times1\times512$ & -- & -- \\ 
  max-pool1/5 & $16\times16/16$ & $2\times2\times128$ & -- & --\\ 
  max-pool2/5 & $8\times8/8$ & $4\times4\times32$ & -- & -- \\ \hline 
  
  fully-conv/6 & -- & $1\times1024$ & $1.51$M & $1.51$M \\ 
  fully-conv/7 & -- & $1\times1024$ & $1.04$M & $1.04$M \\ 
  fully-conv/8 & -- & $1\times40$ & $82$K & $82$K \\ 
  fully-conv/8 & -- & $1\times40$ & $82$K & $82$K\\ 
  pose T/9 & -- & $1\times3$ & $0.1$K & $0.1$K \\  
  pose R/9 & -- & $1\times4$ & $0.1$K & $0.1$K \\ \hline 
  & & & $ \approx 3$M & $346.3$M \\ \hline 
\end{tabular}
\end{center}
\caption{A detailed descriptions of the number of parameters and floating point operations (FLOPs) utilized at different layers in the proposed SPP-Net.}
\label{tab:paramnum}
\end{table}

\begin{table}[!bht]
\setlength{\tabcolsep}{6pt}
\scriptsize 
\begin{center}
  \begin{tabular}{p{4.1cm}|c|c} \hline 
  		 Method & ~$\#$params~ & ~$\#$FLOPs~ \\ \hline  \hline   
  		 SPP-Net (Proposed) & $3$M & $0.35$B  \\ 
  		 Original PoseNet (GoogleNet) \cite{kendall2015posenet} & $8.9$M & $1.6$B \\
  		 Baseline (ResNet50) \cite{laskar2017camera,melekhov2017image} & $26.5$M & $3.8$B \\ 
  		 PoseNet LSTM \cite{weyand2016planet} & $9.0$M & $1.6$B \\  \hline  
  \end{tabular}
\end{center}
\caption{Comparison on the number of parameters and floating point operations (FLOPs).} 
\label{tab:paramcomp}  
\end{table}

\paragraph{Loss function}
We follow~\cite{kendall2017geometric} in the choice of the loss,
\begin{align}
  \cL_\sigma (q, T) & \propto \sigma^{-2}_q\left\lVert q^{\dagger} - q/\|q\| \right\rVert + \sigma^{-2}_T \|T^{\dagger} - T \| \nonumber \\
  &+ \log \sigma_q^2 + \log \sigma_T^2
\label{eq:loss} 
\end{align}
where $q^{\dagger}$ and $T^{\dagger}$ are the ground truth orientation and position of the image, respectively.
We employ unit quaternions to represent 3D rotations.
Note the reprojection error could be a geometrically more meaninful loss function, especially since SPP-Net takes sparse features as input. As also pointed out in~\cite{kendall2017geometric}, we found it difficult to train a network directly using the reprojection loss, hence we rely on~\eqref{eq:loss} instead.

\section{Mining new views}\label{sec:mining}
In this section, we discuss our proposed method for mining synthetic poses and generating synthetic features ``images''.
Before generating synthetic views, we remove all the points in the point-cloud which are either seen in only the test images or observed in fewer than two training images. Also, the observed image indices and the respective feature descriptors for the remaining points corresponding to the test images are removed. Thus, in our preprocessed 3D point cloud there is no information about the test image set. 

Each point $\bX_i$ in the reconstructed 3D point cloud $\cD$ contains the 3D location $(X_i, Y_i, Z_i)$, 
the indices of the images $\cI_{\bX_i}$ where the point $\bX_i$ was observed and the indices of the keypoints in the observed image. Moreover, the positions and the orientations of the images $\cI_{\bX_i}$  are also available. This enables us to synthesize more realistic unobserved views from the 3D point cloud $\cD$. Inspired by the idea of view synthesis in the context of absolute pose estimation~\cite{irschara2009structure,sattler2016large,torii201524} we utilize a similar strategy as described below.

\subsection{Pose set augmentation} \label{sec:poseaug}
\noindent For outdoor datasets, a horizontal plane is robustly fitted to the training camera positions. The synthetic poses are then generated by perturbing each training pose along the detected horizontal plane. Translations and orientations are chosen uniformly within the range of $[-2.5$m$, 2.5$m$]$ and $[-30\degree, 30\degree]$ respectively. The axis of the angular shift is chosen as the normal to the detected horizontal plane.  For indoor datasets the random shifts are given along all the directions uniformly in the interval $[-0.25$m$, 0.25$m$]$ and random orientations $[-30\degree, 30\degree]$ along arbitrary axes. The traversal is performed $50$ times for each of the training pose. The intrinsic camera parameters (focal length, radial distortions) of the synthetic views were chosen to be the same as the training images. Note that in none of the datasets any prior knowledge of the test poses are exploited during the pose augmentation. 


The set of synthetic poses is made more distinctive by removing all the repetitive poses, \eg, poses within $0.1$m location and $1\degree$ orientation of an existing training pose. Furthermore, synthetic poses inside the point-cloud or extremely closed to it (more than $25$ points inside the frustum and within radius $1$m from the camera center) are not useful and realistic, thus have been removed consequently. Note that no optimal placement and orientation strategies are utilized unlike in~\cite{irschara2009structure}. However, we believe that the performance of the network could generalize well if the augmented poses cover the area of the concerned view-points of interest.  The current pose mining produces best results on the benchmarking datasets. 

\subsection{Generate synthetic views}\label{sec:genview}
\noindent For a given camera pose $\cP$, we project the points $\bX$ from the point-cloud $\cD$ in the front of the camera and within the viewing frustum. Not all the points might be relevant here---we keep only those points which ensure \emph{detectability} and  \emph{repeatability} of the descriptors under perspective distortions. Based on that the following selection criterion are made:  \begin{enumerate}[leftmargin=1em,itemsep=1pt,parsep=1pt,topsep=1pt]
\item the relative scale $ s_\bX $ of the projected point of $\bX$ must be greater than $1.25$ and less than $120.0$, 
\item at least one of the original viewing direction must be oriented within $20\degree$ of the current view.  
\end{enumerate}
Note that the scale $s_\bX $ is computed as the relative scale w.r.t.\ the observed image under consideration. 

Once a 3D point is chosen, the feature descriptor $\bbf_i$ is copied corresponding to the nearest observed image. The pixel coordinate $(p_i, q_i)$ of the feature point is computed under the perspective projection of the 3D point $\bX$.  The rotation of the feature descriptor is copied from the chosen observed image. 
At this point we further discard poses, in which the projected point cloud are not sufficiently well spatially distributed in the image: at least four of $4\times 4$ bins arranged over the image have to be non-empty for a sampled pose to be processed further.

 In order to make the synthesized view robust to noise and outliers, we add the following sources of noise:
\begin{itemize}[leftmargin=1em,itemsep=1pt,parsep=1pt,topsep=1pt]
\item Additive Gaussian noise with diagonal co-variance $\Sigma_x$ is added with the feature descriptors $\bbf_i$. The co-variance matrix $\Sigma_x$ is determined based on the descriptors in the training data. 
Further, the projected pixel locations $(p_i, q_i)$ are corrupted by Gaussian noise with variance of $1$ pixel.
\item Outliers comprising of $25\%$ of the total number of projected points are added from randomly chosen pixel locations. The feature descriptors, scale and rotations of the outliers are copied from randomly chosen 3D points. 
\item Outlier keypoints from the training images, that are not utilized for SfM reconstruction, are also projected to the synthetic poses. In this case, we fit an homography through all the inlier points of the synthesized pose and the training pose,  and then use the same homography to project the outliers ($25\%$) to the target synthetic image. This step is omitted if the number of common inliers is less than $50$. 
\end{itemize}
\noindent The above procedure is summarized in algorithm~\ref{algo:algo1}  
\begin{algorithm}
\scriptsize
\algsetup{linenosize=\tiny} 
\SetAlFnt{\tiny\sffamily}
    \SetKwInOut{Input}{Input}
    \SetKwInOut{Output}{Output}
\SetInd{1em}{1em}
  \SetNoFillComment
  \SetCommentSty{scriptsize}
    \Input{3D map $\{\cD\}$, Training feature descriptors $\{\cI\}$, and poses $\{\cP_\cI\}$} 
    \Output{Synthetic pose and feature descriptors $\{\cP, \cF\}$} 
    
    $\{\cP\} \coloneqq \emptyset$; $\{\cP, \cF\} \coloneqq \emptyset$; $i \coloneqq 0$ \\ 
    \ForAll{$\cP \in \{\cP_\cI\}$ }{          
    	\lWhile{$i<50;i$++}{$\{\cP\}$.append $\coloneqq$ augment pose ($\cP$)} 
    } 
    $\{\cP\}$ $\coloneqq$ prune repetitive pose $\{\cP\}$ \tcc*[r]{[see~sec~\ref{sec:poseaug}]}
    \ForAll{$\cP \in \{\cP\}$ }{ 
    	$\cF = \emptyset$;  
    	project $\bX$ into the camera plane $\forall\bX \in \{\cD\}$; \\ 
    	\ForAll{$\bX \in $ viewing frustum of $(\cP)$ }{ 
    		\ForAll{observed images $\cI_\bX$ of the 3D point $\bX$}{ 
		    \lIf{  $s_\bX \notin [1.25, 120]$ }{\emph{continue}} 
		    \lIf{ $\angle(\cP, \cI_\bX) > 20\degree$ }{\emph{continue}} 
		 	$\cF$.append $\coloneqq$ projected pixel location of $\bX$ + scale $s_\bX$ and orientation $\theta$ + descriptors at $\bX$ in $\cI_\bX$ \tcc*[r]{[see~sec~\ref{sec:genview}]} 
		 	
		 	}
		 	}
	\If{ $\cF$ passes the sustainability check}
      {
      $\cH\coloneqq$ homography between the inliers of $\cI$ and $\cF$; \\ 
      $\cF$.append $\coloneqq$ outlier keypoints of $\cI$ projected by $\cH$; \\ 
      $\cF$.perturbation $\big($noise $\&$ outliers$\big)$ \tcc*[r]{[see~sec~\ref{sec:genview}]} 
        $\{\cP, \cF\}$.append $\coloneqq$ $(\cP, \cF)$; 
      }
    }
   
    \KwRet{$\{\cP, \cF\}$} \; 
    \caption{\footnotesize Synthetic pose generation from 3D map }
    \label{algo:algo1}
\end{algorithm}

\section{Experiments}

The proposed SPP-Net is trained on a number of widely used datasets for absolute pose estimation using a Tensorflow~\cite{abadi2015tensorflow} implementation. The loss \eqref{eq:loss} is minimized using ADAM~\cite{kingma2015adam} with a batch size of $300$. The weight decay is set to $10^{-5}$. The network is trained for $400$ epochs with an initial learning rate $0.001$ which is gradually decreased by a factor of $10$ after every $100$ epochs. All the experiments are evaluated on a desktop equipped with a NVIDIA Titan X GPU, where evaluation of the SPP-Net requires about $2$ms of run-time.   The network takes $2-4$hrs to train for a typical dataset and only $36.8$ MB to store the weights.  

\subsection{Validation of the proposed pose augmentation}\label{sec:justification}
\begin{figure}
\centering     
\subfigure[training data only]{\includegraphics[width=0.22\textwidth]{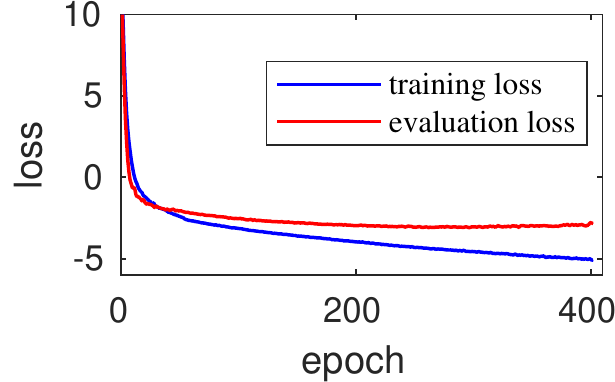}}~
\subfigure[additional synthetic data]{\includegraphics[width=0.22\textwidth]{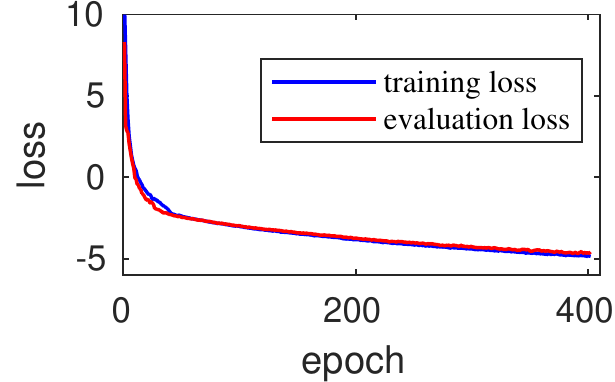}}
\caption{Training and evaluation loss on testing dataset. When the model is trained with the training data only the evaluation loss quickly become stagnant.} 
\label{fig:comparison1}
\end{figure}

\noindent To validate the efficiency of the augmented poses, we conduct an experiment with one of the difficult sequences (heads) of Microsoft's 7-Scenes Dataset~\cite{shotton2013scene}. From the 3D map of training images, we synthetically generate views corresponding to \emph{test poses}. The proposed SPP-Net is then trained on the training images + the  synthetically generated images (sparse feature descriptors) and evaluated on the feature descriptors extracted from the original test images. The generated synthetic test images do not exploit any test image content but the 6 d.o.f.\ poses\footnote{Note that except the current experiment, the synthetic feature descriptors do not incorporate any information of the 6 d.o.f.\ test poses.}.
If the networks is provided only with the training images, it does not generalize well to the test images.
However, after adding synthetic test poses to the training data, the evaluation loss decreases in conjunction with the training loss.
In Figure~\ref{fig:comparison1} we illustrate the training and evaluation loss with and without additional synthetic training data.
These results justify the utilization of our synthetic pose augmentation method.

Using SPP-Net with synthesized poses clearly improves the network's ability to generalize beyond real training poses.
Moreover, SPP-Net can be trained for a target set of poses of interest, which e.g.\ might not be represented in the original training set. This is the main motivation for choosing using sparse features for absolute pose regression.

\begin{table*}\setlength{\tabcolsep}{5pt}
\small
\begin{center}
  \begin{tabular}{p{2.3cm}ccccccc} \hline 
  		& 	Area or & Active Search 				&	Original 							& PoseNet & PoseNet & & SPP-Net (with \\ 
  Scene & Volume & (SIFT) \cite{sattler2017efficient} & PoseNet \cite{kendall2015posenet} &  LSTM \cite{weyand2016planet} & Geo.\ Cost \cite{kendall2017geometric} & SPP-Net & Synthetic data) \\  \hline \hline
  Great Court & $8000$m$^{2}$ & -- & -- &  -- & $6.83 $m, $3.47^{\degree} $ & $13.2 $m, $8.02^{\degree} $ & $5.42 $m, $2.84^{\degree} $  \\  
  King's College & $5600$m$^{2}$ & $0.42$m, $0.55^{\degree}$ & $1.66$m, $4.86^{\degree}$ &  $0.99$m, $3.65^{\degree}$ & $0.88 $m, $1.04^{\degree} $ & $1.91 $m, $2.36^{\degree} $ & $0.74 $m, $0.96^{\degree} $  \\  
  Old Hospital & $2000$m$^{2}$ & $0.44$m, $1.01^{\degree}$ & $2.62$m, $4.90^{\degree}$ &  $1.51$m, $4.29^{\degree}$ & $3.20$m,  $3.29^{\degree} $ & $2.51 $m, $3.74^{\degree} $ & $2.18 $m, $3.92^{\degree} $  \\  
  Shop Facade & $875$m$^{2}$ & $0.12$m, $0.40^{\degree}$ & $1.41$m, $7.18^{\degree}$ &  $1.18$m, $7.44^{\degree}$ & $0.88 $m, $3.78^{\degree} $ & $1.31 $m, $7.82^{\degree} $ & $0.59 $m, $2.53^{\degree} $  \\  
  StMary's Church & $4800$m$^{2}$ & $0.19$m, $0.54^{\degree}$ & $2.45$m, $7.96^{\degree}$ &  $1.52$m, $6.68^{\degree}$ & $1.57 $m, $3.32^{\degree} $ & $3.21 $m, $6.97^{\degree} $ & $1.83 $m, $3.35^{\degree} $  \\  
  Street & $50000$m$^{2}$ & $0.85$m, $0.83^{\degree}$ & -- &  -- & $20.3 $m, $25.5^{\degree} $ & $54.9 $m, $37.2^{\degree} $ & $24.5 $m, $23.8^{\degree} $  \\  \hline  \\ \hline 
 
  Chess & $6$m$^{3}$ & $0.04$m, $1.96^{\degree}$ & $0.32$m, $6.60^{\degree}$ &  $0.24$m, $5.77^{\degree}$ & $0.13 $m, $4.48^{\degree} $ & $0.22 $m, $7.61^{\degree} $ & $0.12 $m, $4.42^{\degree} $  \\  
  Fire & $2.5$m$^{3}$ & $0.03$m, $1.53^{\degree}$ & $0.47$m, $14.0^{\degree}$ &  $0.34$m, $11.9^{\degree}$ & $0.27 $m, $11.3^{\degree} $ & $0.37 $m, $14.1^{\degree} $ & $0.22 $m, $8.84^{\degree} $  \\  
  Heads & $1$m$^{3}$ & $0.02$m, $1.45^{\degree}$ & $0.30$m, $12.2^{\degree}$ &  $0.21$m, $13.7^{\degree}$ & $0.17$m,  $13.0^{\degree} $ & $0.22 $m, $14.6^{\degree} $ & $0.11 $m, $8.33^{\degree} $  \\  
  Office & $7.5$m$^{3}$ & $0.09$m, $3.61^{\degree}$ & $0.48$m, $7.24^{\degree}$ &  $0.30$m, $8.08^{\degree}$ & $0.19 $m, $5.55^{\degree} $ & $0.32 $m, $10.0^{\degree} $ & $0.16 $m, $4.99^{\degree} $  \\  
  Pumpkin & $5$m$^{3}$ & $0.08$m, $3.10^{\degree}$ & $0.49$m, $8.12^{\degree}$ &  $0.33$m, $7.00^{\degree}$ & $0.26 $m, $4.75^{\degree} $ & $0.47 $m, $10.2^{\degree} $ & $0.21 $m, $4.89^{\degree} $  \\  
  Red Kitchen & $18$m$^{3}$ & $0.07$m, $3.37^{\degree}$ & $0.58$m, $7.54^{\degree}$ &  $0.24$m, $5.52^{\degree}$ & $0.23$m, $5.35^{\degree} $ & $0.34 $m, $11.3^{\degree} $ & $0.21 $m, $4.76^{\degree} $  \\   
  Stairs & $7.5$m$^{3}$ & $0.03$m, $2.22^{\degree}$ &  $0.48$m, $13.1^{\degree}$  &   $0.40$m, $13.7^{\degree}$  & $0.35$m, $12.4^{\degree} $ & $0.40 $m, $13.2^{\degree} $ & $0.22 $m, $7.17^{\degree} $  \\  \hline 
 
\end{tabular}
\end{center}
\caption{Median localization results for the Cambridge Landmarks \cite{kendall2015posenet} and
seven Scenes datasets \cite{glocker2013real}. We compare the performance of various training-based algorithms (some baseline entries are copied from \cite{weyand2016planet}). Active search \cite{sattler2017efficient} is a sota traditional geometry based baseline. We demonstrate a notable improvement over the original PoseNet~\cite{kendall2015posenet} and re-projection error proposed in this paper, narrowing the
margin to the geometry based techniques.}
\label{tab:comparison}  
\end{table*}

\begin{figure}
\centering     
{\includegraphics[width=0.22\textwidth]{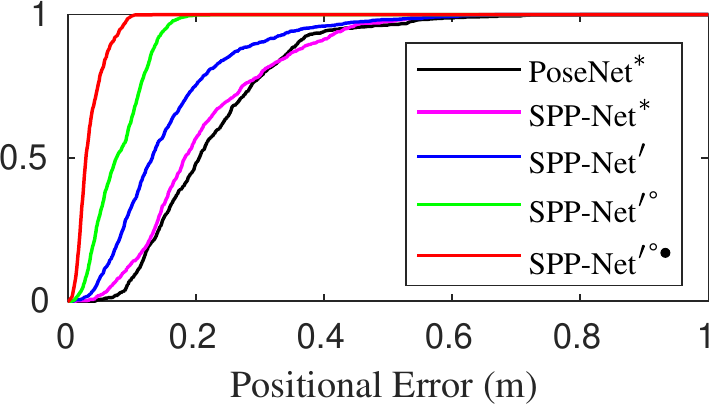}} ~~
{\includegraphics[width=0.22\textwidth]{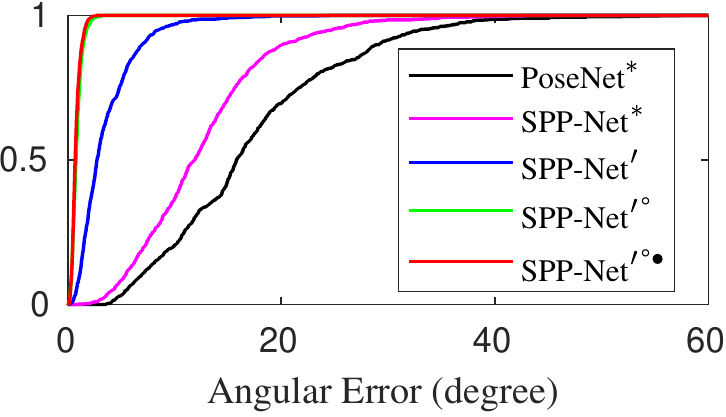}}
\caption{The above figures demonstrate our localization accuracy for both position and orientation as a cumulative histogram of errors for the entire testing set. Where the baselines---Net$^\ast$: trained with the training data only, Net$^\prime$: trained with the clean synthetic data, Net$^{\prime\circ}$: trained with the synthetic data under realistic noise, Net$^{\prime\circ\bullet}$: trained with the synthetic data under realistic noise and outliers.}
\label{fig:comparison2}
\end{figure}

We perform another experiment to validate different steps of the proposed augmentation, where we generate three different sets of synthetic poses with increasing realistic adjustment on each step of the synthetic image generation process. The first set of synthetic poses contains no noise or outliers, the second set is generated with added noise, and the third set is generated with added noise and outliers as described above. Note that all the networks are evaluated on the original sparse test feature descriptors. We also evaluate PoseNet~\cite{kendall2015posenet}, utilizing a tensorflow implementation available online \footnote{\href{https://github.com/kentsommer/keras-posenet}{github.com/kentsommer/keras-posenet}}, trained on the original training images for $800$ epochs. The proposed SPP-Net, trained only on the training images, performs analogously to PoseNet. However, with the added synthetic poses the performance improves immensely with the realistic adjustments as shown in Figure~\ref{fig:comparison2}. Note that since PoseNet uses full image, it cannot easily benefit from augmentation. 

\begin{figure}
\centering     
{\includegraphics[width=0.22\textwidth]{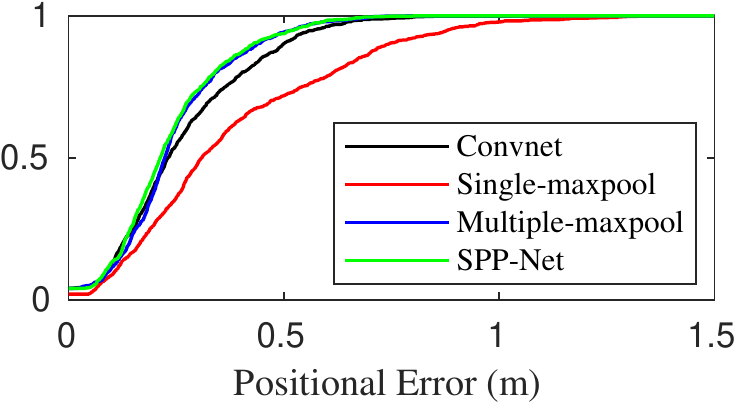}} ~~
{\includegraphics[width=0.22\textwidth]{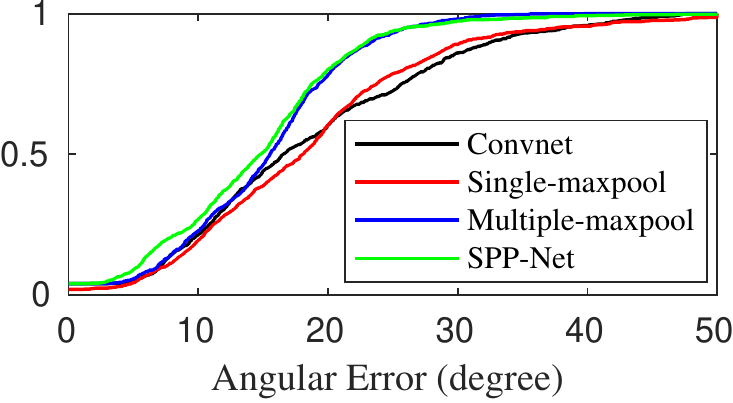}} \\
{\includegraphics[width=0.45\textwidth]{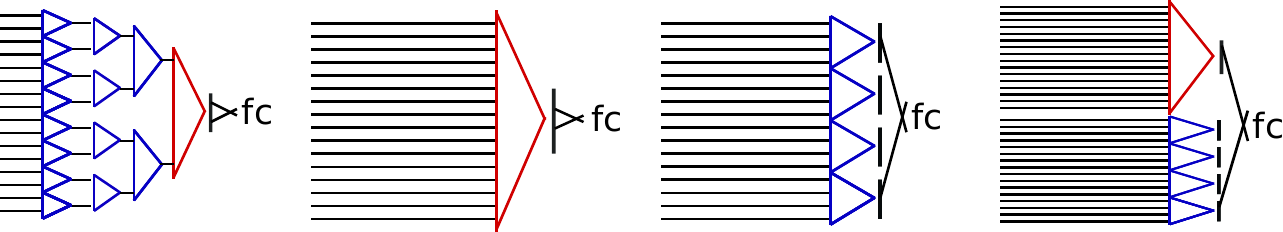}} \\
\scriptsize (a) ConvNet ~~~~(b) Single-maxpool ~~~~ (c) Multiple-maxpool~~~~~~~~(d) SPP-Net ~~~~~

\caption{Top row: the results with different architecture settings--ConvNet is a conventional feed forward network acting on the sorted sparse descriptors. Single-maxpool and Multiple-maxpool are when only a single maxpooling unit at level-0 and multiple maxpooling at level-2 is used.  We observe better performance when we combine those in SPP-Net. Bottom row: 1D representation of different architectures where the convolutions and maxpooling unites are represented by horizontal lines and triangles respectively. The global max-pooling is colored by red and other maxpooling unites are colored by blue.}
\label{fig:comparison3}
\end{figure}

An additional experiment is conducted to validate the architecture of SPP-Net. In this experiment, the SPP-Net is evaluated with the following architecture settings: 
\begin{itemize}[leftmargin=1em,itemsep=1pt,parsep=1pt,topsep=1pt]
\item ConvNet: conventional feed forward network with convolution layers and max-pooling layers are stacked one after another (same number of layers and parameters as SPP-Net) acting on the sorted 2D array of keypoints.
\item Single maxpooling: a single maxpooling layer at level 0, 
\item Multiple maxpooling: one maxpooling layer at level 2,
\item SPP-Net: concatenate responses at three different levels. 
\end{itemize}
In Figure~\ref{fig:comparison3}, we display the results with the different choices of the architectures where we observe best performance with SPP-Net. Note that no synthetic data used in this case.

\subsection{Visualizing leveraged image features}
\noindent
It is instructive to visualize which keypoints extracted in the image are
eventually most relevant to predict the pose parameters. We define the
contribution of a feature to pose prediction as the number of max-pooling
units where the given feature is the winning branch in the max-pooling
step. The higher the contribution, the more prominent is this feature
represented in the following pose regression layers.  Due to the sampling step
to choose a single feature from multiple ones falling into the same cell
(recall Sec.~\ref{sec:overview}), there is an intrinsic randomness in the
network output and in which features are relevant. Hence, in the following we
consider average contribution computed from 100~runs.

\begin{figure}
\centering     
\includegraphics[height=7.3em]{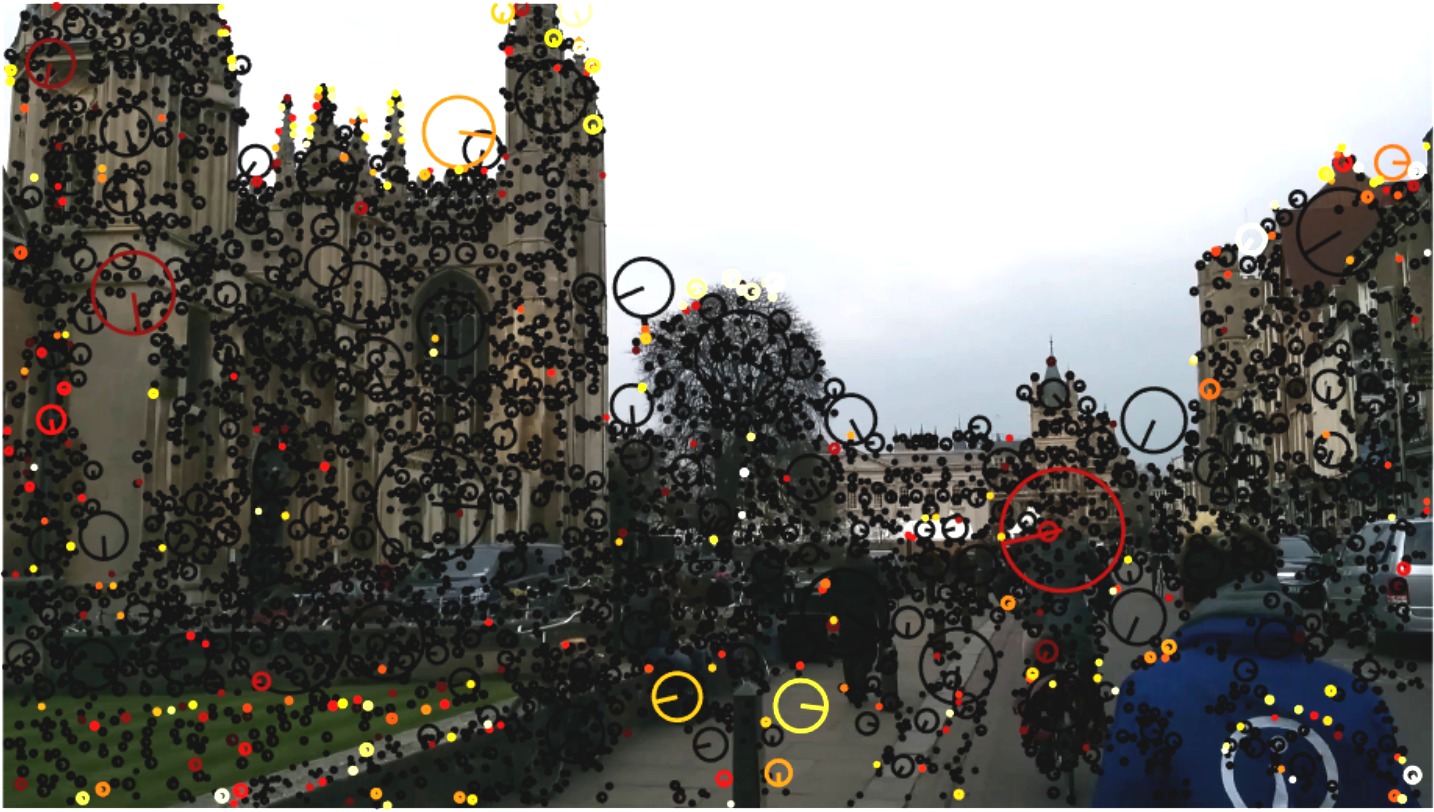}\hspace{1pt}
\includegraphics[height=7.3em]{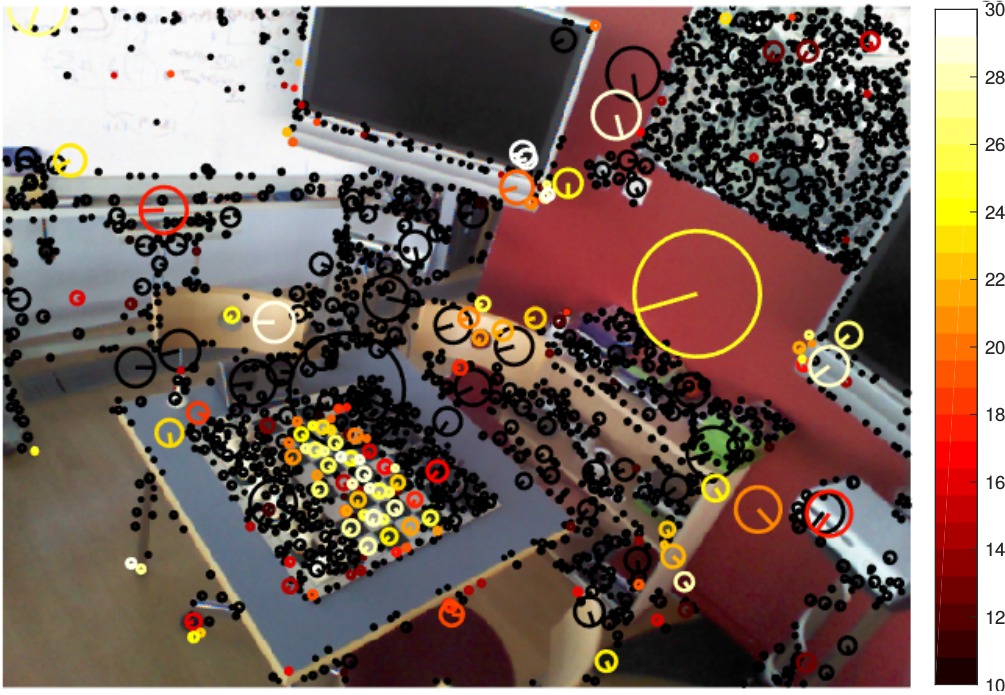} 
\caption{The keypoints for images from the ``Kings College'' and ``Chess''
  sequences are displayed. Highly relevant feature points (with average
  contributions $>10$) are colored (using the ``hot'' colormap) according to
  their contributions to the ensemble layer (see text). In general, feature
  descriptors at larger scales seem to be more relevant. Further, in the
  King's college sequence, keypoints near the building outlines are relatively
  consistently important for pose prediction. The indoor ``Chess'' sequence
  exhibits a mix of features on the unique chess pieces and on the background. }
\label{fig:critical}
\end{figure}

\begin{figure}
\centering     
\begin{tabular}{c@{\hspace{-0.1cm}}c}
{\includegraphics[width=0.22\textwidth]{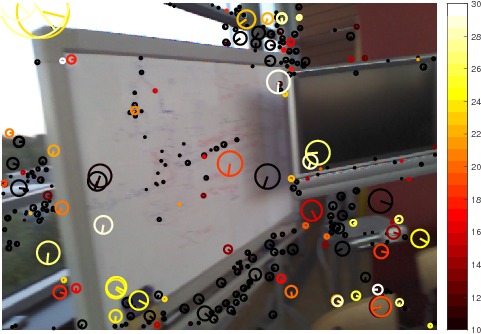}} & 
{\includegraphics[width=0.22\textwidth]{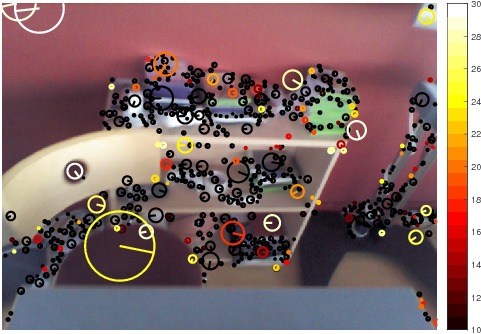}} \\
{\includegraphics[trim=25 10 25 10, clip,width=0.22\textwidth]{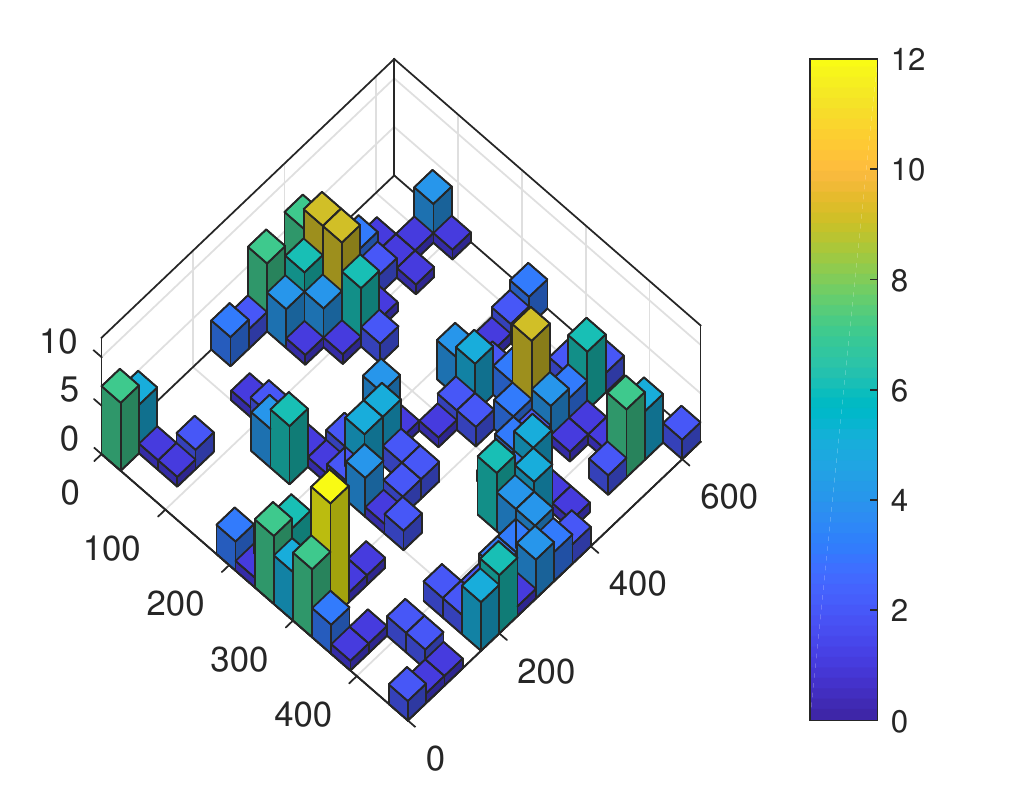}} & 
{\includegraphics[trim=25 10 25 10, clip,width=0.22\textwidth]{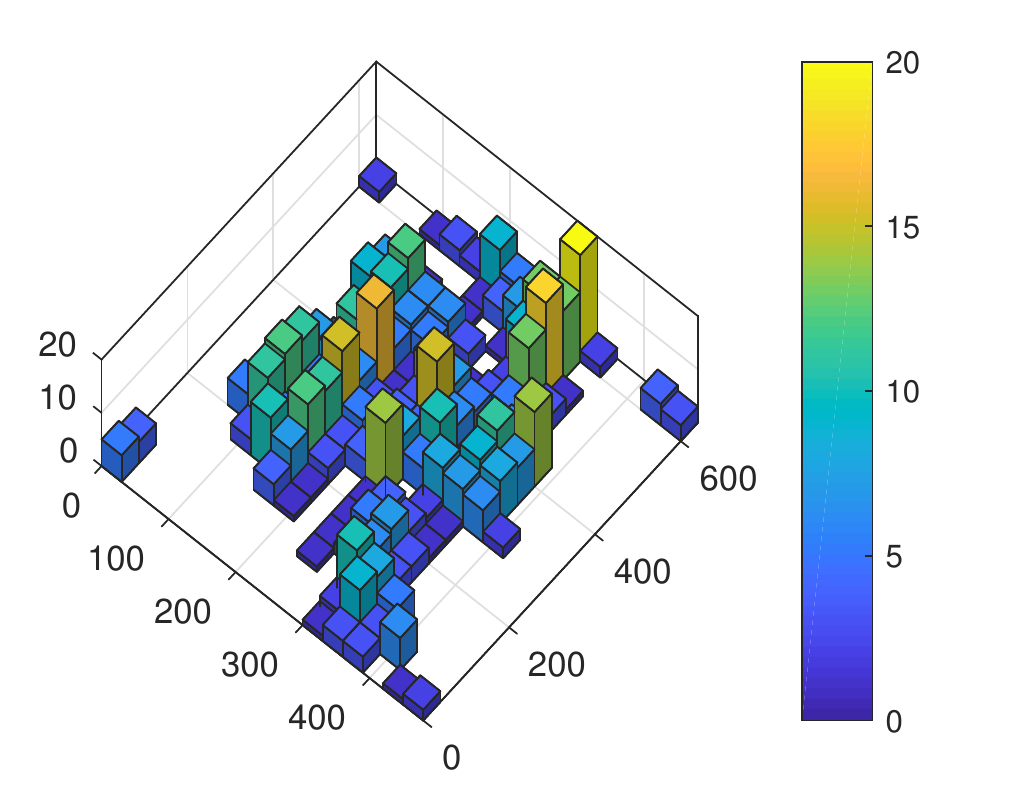}} \\
(a) Angular Error $=12.71\degree$,~~ & ~~(b) Angular Error $=6.52\degree$,   \\
~~Positional Error $=0.36$m &~~Positional Error $=0.21$m 
\end{tabular}
\caption{A pair of typical test images of the ``chess'' sequences are displayed along with the histograms where $56.2\%$ and $51.8\%$ cells of the $16\times 16$ grids are empty. Although, our method predicts the pose successfully, it is bit worse than the median prediction mentioned in table~\ref{tab:comparison}.}
\label{fig:comparison33}
\end{figure}

In Fig~\ref{fig:critical} we display the most contributing feature points for
two complementary scenes. For outdoor environments many features relevant for
pose prediction cluster near the skyline induced by building, and for indoor
scenarios one generally observes a mix between distinctive small-scale
features and background features at a larger keypoint scale.  Further, in
Fig.~\ref{fig:comparison33}, we display a pair of images where more than
$50\%$ of bins (cells) are empty yet SPP-Net successfully estimates the
pose. This indicates that SPP-Net shows robustness to unevenly distributed
image features.

A video ({\tt chess.mov}\footnote{\href{https://youtu.be/Fuv18OMaTnk}{https://youtu.be/Fuv18OMaTnk}}) is uploaded that visualizes the ``Chess'' sequence with overlaid features. The relevance of features is determined and visualized as in Fig.~\ref{fig:critical}. A relatively small and also temporally coherent set of salient features is chosen by SPP-Net for pose estimation. 

%
%

\subsection{Benchmarking localization accuracy}
\noindent {\bf Baseline Methods}
We compare the proposed SPP-Net against the following baselines: 
\begin{itemize}[leftmargin=1em,itemsep=1pt,parsep=1pt,topsep=1pt]
\item Active Search \cite{sattler2017efficient}: This is a direct feature-based approach where the feature descriptors are matched across the 3D point-cloud and the absolute pose is estimated robustly utilizing the P3P algorithm.     
\item Original PoseNet \cite{kendall2015posenet}: The first convnet-based method where the last soft-max classification layer of GoogleNet is replaced by the fully connected regression layers.  
\item PoseNet LSTM \cite{weyand2016planet}     :  Similar as above, but multiple LSTM units were utilized to the convnet features followed by a regression layers. 
\item PoseNet Geometric Cost \cite{kendall2017geometric} (PoseNet2): The network is trained with the same loss function as ours and fine-tuned with the re-projection cost. 
\item Proposed SPP-Net trained without augmented posses is also included as baseline. 
\end{itemize}   

\noindent {\bf Cambridge Landmarks Datasets}
The datasets \cite{kendall2015posenet} provides a labeled set of image sequences of different outdoor scenes where the ground-truth poses were obtained by utilizing VisualSFM~\cite{wu2011visualsfm}. The datasets also provide the SfM ``reconstruction''($.nvm$) files containing the 3D point-cloud and the 2D-3D assignments required by our pose augmentation.
Creation of synthetic images takes approximately two hours for a typical dataset (per sequence). The SPP-Net is trained on the augmented (training and synthesized) dataset. 
The results are displayed in Table~\ref{tab:comparison}. The SPP-Net produces comparable results with the original PoseNet and comparable with PoseNet2~\cite{kendall2017geometric} once trained on an augmented dataset, however, SPP-Net is more lightweight, much faster and not required to be pre-trained on a larger datasets. Note that the proposed network is of limited size, increasing the size of the network (and essentially size of the augmented poses) will further improve the performance. 

\noindent {\bf Seven Scene Datasets}
The Microsoft 7-Scenes Dataset~\cite{shotton2013scene} consists of texture-less RGB-D images of seven different indoor scenes. The scenes were captured with a Kinect device  and reference poses were obtained using KincetFusion~\cite{izadi2011kinectfusion}. The dataset was originally constructed for RGB-D based localization as the scene is extremely challenging for direct approach with the texture-less feature descriptors.  

The 3D map and the feature descriptors are not provided with the datasets required by the proposed augmented pose generation technique. Thus, we reconstructed the 3D point cloud from scratch using toolboxes such as VisualSFM~\cite{wu2011visualsfm} and COLMAP~\cite{schonberger2015single}.
We register the SfM camera poses to the KincetFusion reference poses by a similarity transformation, and the same transformation is used to register the 3D points w.r.t.\ the reference poses.
The points obtained by SfM reconstruction are refined using the reference poses.
Note that as the reference poses are rather noisy, hence the similarity transformation does not relate SfM poses and reference poses well in all scenes. 
In particular, we observed good results on ``Stairs'', ``Heads'' and ``Fire'' sequences as the similarity transformation is a good fit for these scenes as shown in Table~\ref{tab:comparison}. Overall we have obtained very competitive results in this dataset.

\subsection{Varying network size}
\noindent
This experiments aims to determine the sensitivity of the SPP-Net architecture
to the number of network parameters. We consider two modifications for the network size:
\begin{itemize}[leftmargin=1em,itemsep=1pt,parsep=1pt,topsep=1pt]
\item half the number of feature channels used in convolutional and fully connected layers of SPP-Net,
\item conversely, double the number of all feature channels and channels in the fully connected layers.  
\end{itemize}
As a result we have about one fourth and $4\times$ number of parameters,
respectively, compared to our standard SPP-Net. The above networks are trained
on the augmented poses of the seven Scenes datasets. The results are displayed
in Table~\ref{tab:comparisonsize} and indicate, that the performance of the
smaller network is degrading relatively gracefully, whereas the larger network
offers insignificant gains (and it seems to show some signs of overfitting).

In Table~\ref{tab:comparisonsize}, we display the results on Cambridge Landmark Datasets \cite{kendall2015posenet} where  
we observe similar performance as above. It improves the performance with the size of the network for most of the sequence, except the sequence ``Shop Facade''. Again, we believe that in this case the larger network starts to overfit on this smaller dataset.  

\begin{table}\setlength{\tabcolsep}{5pt}
\small
\begin{center}
  \begin{tabular}{p{1.3cm}cccc} \hline 
  		 & SPP-Net ($0.25\times$) & SPP-Net & SPP-Net ($4\times$)\\  \hline \hline
  Chess  &  $0.15 $m, $4.89^{\degree} $    &  $0.12 $m, $4.42^{\degree} $ & $0.10 $m, $3.36^{\degree} $ \\  
  Fire  &   $0.28 $m, $12.4^{\degree} $   &  $0.22 $m, $8.84^{\degree} $   &    $0.21 $m, $8.35^{\degree} $  \\  
  Heads  & $0.14 $m, $10.7^{\degree} $   &  $0.11 $m, $8.33^{\degree} $ &    $0.11 $m, $8.06^{\degree} $  \\  
  Office  &  $0.19 $m, $6.15^{\degree} $    & $0.16 $m, $4.99^{\degree} $  &   $0.13 $m, $4.07^{\degree}$ \\  
  Pumpkin &  $0.34 $m, $8.47^{\degree} $    & $0.21 $m, $4.89^{\degree} $  &   $0.20 $m, $5.35^{\degree} $   \\
  Red Kit. & $0.26 $m, $5.16^{\degree} $ & $0.21 $m, $4.76^{\degree} $ &  $0.22 $m, $5.29^{\degree} $ \\   
  Stairs & $0.25 $m, $7.38^{\degree} $   & $0.22 $m, $7.17^{\degree} $ &  $0.20 $m, $7.25^{\degree} $ 
\\  \hline 
 
\end{tabular}
\end{center}
\caption{Evaluation of SPP-Net with varying number of parameters on seven Scenes datasets.}
\label{tab:comparisonsize}  
\end{table}

\begin{table}[!htbp]\setlength{\tabcolsep}{3pt}
\small 
\begin{center}
  \begin{tabular}{p{1.7cm}ccccc} \hline 
  		&   Ours   & Ours   & Ours    \\ 
  		&  SPP-Net ($0.25\times$) & SPP-Net & SPP-Net ($4\times$) \\ 
\hline \hline
  Great Court  &  $7.58 $m, $5.91^{\degree} $ & $5.42 $m, $2.84^{\degree} $  & $5.48 $m, $2.77^{\degree} $  \\  
  King's Coll.    & $1.41 $m, $2.02^{\degree} $ & $0.74 $m, $0.96^{\degree} $ & $0.83 $m, $1.01^{\degree} $  \\  
  Old Hosp. &  $2.06 $m, $3.91^{\degree} $ & $2.18 $m, $3.92^{\degree} $ & $1.83 $m, $3.25^{\degree} $  \\  
  Shop Facade   &  $0.87 $m, $3.36^{\degree} $ & $0.59 $m, $2.53^{\degree} $ & $0.64 $m, $3.05^{\degree} $\\  
  StMary's Ch.   &  $2.26 $m, $6.46^{\degree} $ & $1.83 $m, $3.35^{\degree} $  & $1.62 $m, $3.42^{\degree} $ \\  
  Street  &  $33.9 $m, $31.2^{\degree} $ & $24.5 $m, $23.8^{\degree} $  & $17.5 $m, $20.2^{\degree} $ \\  \hline  
 
\end{tabular}
\end{center}
\caption{Evaluation of SPP-Net with varying number of parameters on Cambridge Landmark datasets \cite{kendall2015posenet}.}
\label{tab:comparisonsize2}  
\end{table}

\section{Conclusion}

In this work we presented a deep learning architecture for pose prediction
able to learn from real and synthesized views. Thus, pose regression can be
trained for any region in the space of all poses using a virtually unlimited
amount of (synthetic) training data. Our proposed method to create synthetic
data aims to be sufficiently realistic by using an underlying 3D point cloud
and an outlier and noise generation model. We performed a number of numerical
experiments to validate our architecture and the proposed augmentation
procedure, and we achieve state-of-the-art results on benchmark datasets for
pose regression.

 

{
\small
\bibliographystyle{ieee}
\bibliography{sample}

\begin{thebibliography}{10}\itemsep=-1pt

\bibitem{abadi2015tensorflow}
M.~Abadi, A.~Agarwal, P.~Barham, E.~Brevdo, Z.~Chen, C.~Citro, G.~Corrado,
  A.~Davis, J.~Dean, M.~Devin, et~al.
\newblock Tensorflow: Large-scale machine learning on heterogeneous distributed
  systems.
\newblock 2015.

\bibitem{angeli2008fast}
A.~Angeli, D.~Filliat, S.~Doncieux, and J.-A. Meyer.
\newblock Fast and incremental method for loop-closure detection using bags of
  visual words.
\newblock {\em IEEE Transactions on Robotics}, 24(5):1027--1037, 2008.

\bibitem{clement2017train}
L.~Clement and J.~Kelly.
\newblock How to train a cat: Learning canonical appearance transformations for
  robust direct localization under illumination change.
\newblock {\em arXiv preprint arXiv:1709.03009}, 2017.

\bibitem{cummins2008fabmap}
M.~Cummins and P.~Newman.
\newblock Fab-map: Probabilistic localization and mapping in the space of
  appearance.
\newblock {\em The International Journal of Robotics Research}, 27(6):647--665,
  2008.

\bibitem{dosovitskiy2016inverting}
A.~Dosovitskiy and T.~Brox.
\newblock Inverting visual representations with convolutional networks.
\newblock In {\em Proc. CVPR}, pages 4829--4837, 2016.

\bibitem{glocker2013real}
B.~Glocker, S.~Izadi, J.~Shotton, and A.~Criminisi.
\newblock Real-time rgb-d camera relocalization.
\newblock In {\em Mixed and Augmented Reality (ISMAR), 2013 IEEE International
  Symposium on}, pages 173--179. IEEE, 2013.

\bibitem{he2014spatial}
K.~He, X.~Zhang, S.~Ren, and J.~Sun.
\newblock Spatial pyramid pooling in deep convolutional networks for visual
  recognition.
\newblock In {\em Proc. ECCV}, pages 346--361. Springer, 2014.

\bibitem{irschara2009structure}
A.~Irschara, C.~Zach, J.-M. Frahm, and H.~Bischof.
\newblock From structure-from-motion point clouds to fast location recognition.
\newblock In {\em Proc. CVPR}, pages 2599--2606. IEEE, 2009.

\bibitem{izadi2011kinectfusion}
S.~Izadi, D.~Kim, O.~Hilliges, D.~Molyneaux, R.~Newcombe, P.~Kohli, J.~Shotton,
  S.~Hodges, D.~Freeman, A.~Davison, et~al.
\newblock Kinectfusion: real-time 3d reconstruction and interaction using a
  moving depth camera.
\newblock In {\em Proceedings of the 24th annual ACM symposium on User
  interface software and technology}, pages 559--568. ACM, 2011.

\bibitem{jegou2008hamming}
H.~Jegou, M.~Douze, and C.~Schmid.
\newblock Hamming embedding and weak geometric consistency for large scale
  image search.
\newblock {\em Proc. ECCV}, pages 304--317, 2008.

\bibitem{jegou2010improving}
H.~J{\'e}gou, M.~Douze, and C.~Schmid.
\newblock Improving bag-of-features for large scale image search.
\newblock {\em International journal of computer vision}, 87(3):316--336, 2010.

\bibitem{kendall2017geometric}
A.~Kendall and R.~Cipolla.
\newblock Geometric loss functions for camera pose regression with deep
  learning.
\newblock In {\em Proc. CVPR}, 2017.

\bibitem{kendall2015posenet}
A.~Kendall, M.~Grimes, and R.~Cipolla.
\newblock Posenet: A convolutional network for real-time 6-dof camera
  relocalization.
\newblock In {\em Proc. ICCV}, pages 2938--2946, 2015.

\bibitem{kingma2015adam}
D.~Kingma and J.~Ba.
\newblock Adam: A method for stochastic optimization.
\newblock In {\em International Conference on Learning Representations}, 2015.

\bibitem{laskar2017camera}
Z.~Laskar, I.~Melekhov, S.~Kalia, and J.~Kannala.
\newblock Camera relocalization by computing pairwise relative poses using
  convolutional neural network.
\newblock {\em arXiv preprint arXiv:1707.09733}, 2017.

\bibitem{li2010location}
Y.~Li, N.~Snavely, and D.~P. Huttenlocher.
\newblock Location recognition using prioritized feature matching.
\newblock In {\em Proc. ECCV}, pages 791--804. Springer, 2010.

\bibitem{li2016worldwide}
Y.~Li, N.~Snavely, D.~P. Huttenlocher, and P.~Fua.
\newblock Worldwide pose estimation using 3d point clouds.
\newblock In {\em Large-Scale Visual Geo-Localization}, pages 147--163.
  Springer, 2016.

\bibitem{lowe2004distinctive}
D.~Lowe.
\newblock Distinctive image features from scale-invariant keypoints.
\newblock {\em IJCV}, 60(2):91--110, 2004.

\bibitem{melekhov2017image}
I.~Melekhov, J.~Ylioinas, J.~Kannala, and E.~Rahtu.
\newblock Image-based localization using hourglass networks.
\newblock {\em arXiv preprint arXiv:1703.07971}, 2017.

\bibitem{newman2005slam}
P.~Newman and K.~Ho.
\newblock Slam-loop closing with visually salient features.
\newblock In {\em Proc. ICRA}, pages 635--642. IEEE, 2005.

\bibitem{nister2006scalable}
D.~Nister and H.~Stewenius.
\newblock Scalable recognition with a vocabulary tree.
\newblock In {\em Proc. CVPR}, volume~2, pages 2161--2168. Ieee, 2006.

\bibitem{perronnin2010large}
F.~Perronnin, Y.~Liu, J.~S{\'a}nchez, and H.~Poirier.
\newblock Large-scale image retrieval with compressed fisher vectors.
\newblock In {\em Proc. CVPR}, pages 3384--3391. IEEE, 2010.

\bibitem{philbin2007object}
J.~Philbin, O.~Chum, M.~Isard, J.~Sivic, and A.~Zisserman.
\newblock Object retrieval with large vocabularies and fast spatial matching.
\newblock In {\em Proc. CVPR}, pages 1--8. IEEE, 2007.

\bibitem{qi2016pointnet}
C.~R. Qi, H.~Su, K.~Mo, and L.~J. Guibas.
\newblock Pointnet: Deep learning on point sets for 3d classification and
  segmentation.
\newblock {\em arXiv preprint arXiv:1612.00593}, 2016.

\bibitem{sattler2016large}
T.~Sattler, M.~Havlena, K.~Schindler, and M.~Pollefeys.
\newblock Large-scale location recognition and the geometric burstiness
  problem.
\newblock In {\em Proc. CVPR}, pages 1582--1590, 2016.

\bibitem{sattler2011fast}
T.~Sattler, B.~Leibe, and L.~Kobbelt.
\newblock Fast image-based localization using direct 2d-to-3d matching.
\newblock In {\em Proc. ICCV}, pages 667--674. IEEE, 2011.

\bibitem{sattler2017efficient}
T.~Sattler, B.~Leibe, and L.~Kobbelt.
\newblock Efficient \& effective prioritized matching for large-scale
  image-based localization.
\newblock {\em IEEE Trans. on Pattern Anal. and Mach. Intell.},
  39(9):1744--1756, 2017.

\bibitem{schindler2007cityscale}
G.~Schindler, M.~Brown, and R.~Szelisk.
\newblock City-scale location recognition.
\newblock In {\em Proc. CVPR}, 2007.

\bibitem{schonberger2015single}
J.~L. Schonberger, F.~Radenovic, O.~Chum, and J.-M. Frahm.
\newblock From single image query to detailed 3d reconstruction.
\newblock In {\em Proc. CVPR}, pages 5126--5134, 2015.

\bibitem{shotton2013scene}
J.~Shotton, B.~Glocker, C.~Zach, S.~Izadi, A.~Criminisi, and A.~Fitzgibbon.
\newblock Scene coordinate regression forests for camera relocalization in
  rgb-d images.
\newblock In {\em Proc. CVPR}, pages 2930--2937, 2013.

\bibitem{simonyan2015verydeep}
K.~Simonyan and A.~Zisserman.
\newblock Very deep convolutional networks for large-scale image recognition.
\newblock In {\em Proc. ICLR}, 2015.

\bibitem{szegedy2015going}
C.~Szegedy, W.~Liu, Y.~Jia, P.~Sermanet, S.~Reed, D.~Anguelov, D.~Erhan,
  V.~Vanhoucke, and A.~Rabinovich.
\newblock Going deeper with convolutions.
\newblock In {\em Proc. CVPR}, pages 1--9, 2015.

\bibitem{tolias2015particular}
G.~Tolias, R.~Sicre, and H.~J{\'e}gou.
\newblock Particular object retrieval with integral max-pooling of cnn
  activations.
\newblock {\em arXiv preprint arXiv:1511.05879}, 2015.

\bibitem{torii201524}
A.~Torii, R.~Arandjelovic, J.~Sivic, M.~Okutomi, and T.~Pajdla.
\newblock 24/7 place recognition by view synthesis.
\newblock In {\em Proc. CVPR}, pages 1808--1817, 2015.

\bibitem{torralba2008small}
A.~Torralba, R.~Fergus, and Y.~Weiss.
\newblock Small codes and large image databases for recognition.
\newblock In {\em Proc. CVPR}, pages 1--8. IEEE, 2008.

\bibitem{walch2017image}
F.~Walch, C.~Hazirbas, L.~Leal-Taix{\'e}, T.~Sattler, S.~Hilsenbeck, and
  D.~Cremers.
\newblock Image-based localization with spatial lstms.
\newblock In {\em Proc. ICCV}, 2017.

\bibitem{weiss2009spectral}
Y.~Weiss, A.~Torralba, and R.~Fergus.
\newblock Spectral hashing.
\newblock In {\em Proc. NIPS}, pages 1753--1760, 2009.

\bibitem{weyand2016planet}
T.~Weyand, I.~Kostrikov, and J.~Philbin.
\newblock Planet-photo geolocation with convolutional neural networks.
\newblock In {\em Proc. ECCV}, pages 37--55. Springer, 2016.

\bibitem{wu2011visualsfm}
C.~Wu et~al.
\newblock Visualsfm: A visual structure from motion system.
\newblock In {\em ~}.

\bibitem{zhang2006image}
W.~Zhang and J.~Kosecka.
\newblock Image based localization in urban environments.
\newblock In {\em 3D Data Processing, Visualization, and Transmission, Third
  International Symposium on}, pages 33--40. IEEE, 2006.

\end{thebibliography}
}

\end{document}